\documentclass[a4paper, 10pt, conference]{ieeeconf}      
\usepackage{FG2017}

\FGfinalcopy 
\usepackage{booktabs}
\usepackage{url}
\usepackage{graphicx}
\usepackage{amsmath}
\usepackage{amssymb}
\usepackage{epsfig}
\usepackage{epstopdf}
\usepackage{cite}
\usepackage{gensymb}
\usepackage{url}
\usepackage[bottom]{footmisc}
\usepackage{multirow} 
\usepackage{color}
\usepackage{subfig,graphicx,amsmath,amssymb}
\usepackage{breqn}
\usepackage{caption}
\usepackage{tabularx}
\usepackage[english]{babel}

\def\Vect#1{{\boldsymbol{#1}}}
\def\Mat#1{{\boldsymbol{#1}}}
\renewcommand{\Vect}[1]{\boldsymbol{\mathbf{#1}}}

\renewcommand{\Mat}[1]{\boldsymbol{\mathbf{#1}}}

\IEEEoverridecommandlockouts                              
\title{\LARGE \bf
KEPLER: Keypoint and Pose Estimation of Unconstrained Faces by Learning Efficient H-CNN Regressors 
}

\author{\parbox{16cm}{\centering
    {\large Amit Kumar, Azadeh Alavi and Rama Chellappa}\\
    {\normalsize
     Department of Electrical and Computer Engineering, CFAR and UMIACS\\
     University of Maryland-College Park,USA\\
     \{akumar14,azadeh,rama\}@umiacs.umd.edu
    }}
}

\begin{document}

\ifFGfinal
\thispagestyle{empty}
\pagestyle{empty}
\else
\author{Anonymous FG 2015 submission\\-- DO NOT DISTRIBUTE --\\}
\pagestyle{plain}
\fi
\maketitle

\begin{abstract}

Keypoint detection is one of the most important pre-processing steps in tasks such as face modeling, recognition and verification. In this paper, we present an iterative method for Keypoint Estimation and Pose prediction of unconstrained faces by Learning Efficient H-CNN Regressors (KEPLER) for addressing the face alignment problem. Recent state of the art methods have shown improvements in face keypoint detection by employing Convolution Neural Networks (CNNs). Although a simple feed forward neural network can learn the mapping between input and output spaces, it cannot learn the inherent structural dependencies. We present a novel architecture called H-CNN (Heatmap-CNN) which captures structured global and local features and thus favors accurate keypoint detecion. H-CNN is jointly trained on the visibility, fiducials and 3D-pose of the face. As the iterations proceed, the error decreases making the gradients small and thus requiring efficient training of DCNNs to mitigate this. KEPLER performs global corrections in pose and fiducials for the first four iterations followed by local corrections in a subsequent stage. As a by-product, KEPLER also provides 3D pose (pitch, yaw and roll) of the face accurately. In this paper, we show that without using any 3D information, KEPLER outperforms state of the art methods for alignment on challenging datasets such as AFW\cite{AFW_dataset_CVPR2012} and AFLW\cite{tugraz:icg:lrs:koestinger11b}.

\end{abstract}
\section{INTRODUCTION}
Keypoint detection on unconstrained  faces is one of the most studied topics in the past decade, as accurate localization of fiducials is a vital pre-processing task for variety of applications. In the last five years, keypoint localization using Convolution Neural Networks (CNN) has received great attention from computer vision researchers. This is mainly due to the availability of large scale annotated unconstrained face datasets such as AFLW\cite{tugraz:icg:lrs:koestinger11b}. 
Works such as\cite{DBLP:journals/corr/ZeilerF13} have hypothesized that as the network gets deeper more abstract information such as identity, pose, attributes are retained while immediate local features are lost. However, various methods \cite{Sun:2013:DCN:2514950.2516090},\cite{CFAN}, and \cite{Zhu_2015_CVPR} directly use CNNs as regressors or use deep features from CNNs to design regressors for predicting keypoints. 

On the other hand, an earlier method of Explicit Shape Regression (ESR) proposed by Cao et al.\cite{DBLP:journals/ijcv/CaoWWS14} achieved superior results by introducing the important concept of non-parametric shape regression for facial keypoint localization. Following \cite{DBLP:journals/ijcv/CaoWWS14}, unconstrained face alignment received great deal of attention and many of its variants \cite{7299048},\cite{DBLP:conf/cvpr/RenCW014},\cite{kazemi2014one},\cite{Sun:2013:DCN:2514950.2516090},\cite{DBLP:journals/corr/KumarRPC16} were published later, using a variety of features producing incremental improvements over \cite{DBLP:journals/ijcv/CaoWWS14}. However, they are all limited by the fixed number of points on the face. In real life applications, there are more challenging datasets such as IJBA\cite{F_2015_CVPR} and AFW\cite{AFW_dataset_CVPR2012}, which do not always have 68 or 49 fixed points mainly due to occlusion or pose variations. As alternatives, researchers moved towards more sophisticated techniques, incorporating 3D shape models\cite{DBLP:journals/corr/ZhuLLSL15},\cite{lfa3d},\cite{pifa}, domain learning\cite{Zhu_2016_CVPR}, recurrent autoencoder-decoder\cite{recdec} and many others. However, one question still remains unanswered: Can cascaded shape regression be applied for an arbitrary face with no prior knowledge ?  

\begin{figure}[t]
\centering
\includegraphics[height=1.7cm,width=8.7cm]{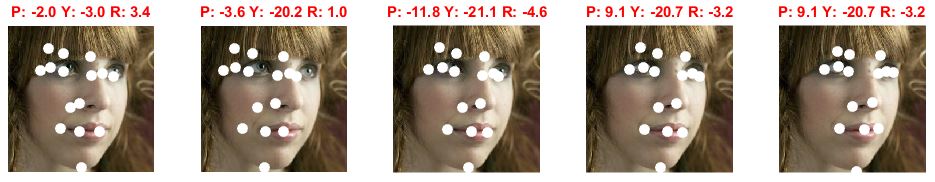}\\
\includegraphics[height=1.7cm,width=8.7cm]{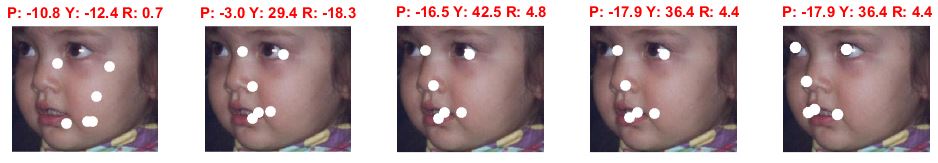}
\caption{\small Sample results generated by the proposed method. White dots represent the location of keypoints after each iteration. The first row shows an image from the AFLW dataset. The points move at subpixel level after fourth iteration. The second row is a sample image from the AFW dataset, which shows how the last stage of error correction can effectively mitigate the inconsistency of the bounding box across datasets. The numbers in red are the predicted 3D pose P:Pitch Y:Yaw R:Roll } 
\label{fig:sample}
\end{figure}
\begin{figure}[htp!]
\centering
\includegraphics[height=5.25cm,width=5.25cm]{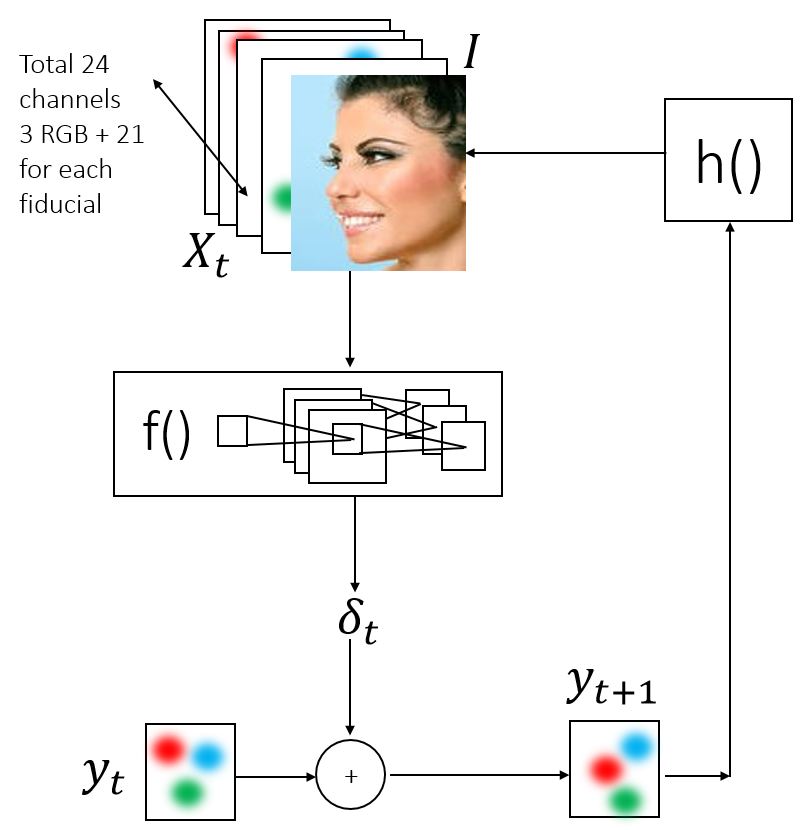}
\caption{\small Overview of the architecture of KEPLER. The function \textit{f()} predicts visibility, pose and the corrections for the next stage. The representation function \textit{h()} forms the input representation for the next iteration.}
\label{fig:method}
\end{figure}

The motivation for this work is to adapt cascaded regression for predicting landmarks of arbitrary faces, while taking advantage of CNNs. We transform the cascaded regression formulation into an iterative scheme for arbitrary faces. In each iteration the regressor predicts the increment for the next stage jointly for all the points while maintaining the shape constraint. As by-products of KEPLER, we get the visibility confidence of each keypoint and 3D pose (pitch, yaw and roll) for the face image. 
The main contributions of this paper are: 
\begin{itemize}
\item We design a novel GoogLenet-based\cite{c2} architecture with a channel inception module which pools features from intermediate layers and concatenates them similar to inception module. We call the proposed architecture \textit{Channeled Inception} in the rest of the paper. This network is used in all the stages of KEPLER.
\item Inspired by \cite{c1}, we present an iterative method for estimating the face landmarks using the fixed point consolidation scheme inspired by \cite{c1}. We observe that estimating landmarks on a face is more challenging than estimating keypoints on a human body. 
The overview of the pipeline is shown in Figure \ref{fig:method}.
\item After each iteration, the error from ground-truth decreases, making the gradient smaller and hence different training policies are employed in every stage for the efficient training of H-CNN.
\item We evaluate the performance of our keypoint estimation method on challenging datasets such as AFLW and AFW, which include faces in diverse poses and expressions. We also introduce a new protocol for evaluating the facial keypoint localization scheme on the AFLW dataset which is more challenging and usually left out while evaluating unconstrained face alignment methods. 
\end{itemize}
The rest of the paper is organized as follows. Section II reviews closely related works. Section III presents the proposed method in detail. Section IV describes the experiments and comparisons, which are then followed by conclusions and suggestions for future works in section V. 

\section{RELATED WORK}
Following \cite{DBLP:journals/ijcv/CaoWWS14}, we classify previous works on face alignment into two basic categories. \\ \\
\textit{\bf{Part-Based Deformable models:}} These methods perform alignment by maximizing the confidence of part locations in a given input image. One of the major works in this category was done by Zhu and Ramanan\cite{AFW_dataset_CVPR2012}, where they used a part-based model for face detection, pose estimation and landmark localization assuming the face shape to be a tree structure.\cite{Asthana:2013:RDR:2514950.2516059} by Asthana et al., learned a dictionary of probability response maps followed by linear regression in a Constrained Local Model (CLM) framework. Hsu et al.\cite{7410796} extended the mixture of tree model \cite{AFW_dataset_CVPR2012} to achieve better accuracy and efficiency. However, their method again assumes face shape to be a tree structure, enforcing  strong  constraints  specific to shape  variations. \\ 
\textit{\bf{Regression-based approaches:}} Since face alignment is naturally a regression problem, a multitude of regression-based approaches has been proposed in recent years. Methods reported in \cite{conf/eccv/LiangXWS08},\cite{DBLP:journals/ijcv/CaoWWS14},\cite{Zhu_2015_CVPR} are based on learning a regression model that directly maps image appearances to target outputs. However, these methods along with methods from \cite{antonakos2016adaptive},\cite{trigeorgis2016mnemonic},\cite{antonakos2015active},\cite{akshay_wild},\cite{6909635} and \cite{6619290} were mostly evaluated either in a lab setting or on face images where all the facial keypoints are visible. Wu et al.\cite{7410774} proposed an  occlusion-robust cascaded regressor to handle occlusion.  Xiong et al.\cite{global2015xiong} pointed out that standard cascaded regression approaches such as Supervised Descent Method (SDM)\cite{XiongD13} tend to average conflicting gradient directions resulting in reduced performance. Hence,\cite{global2015xiong} suggested domain dependent descent maps. Inspired by this, Cascade Compositional Learning (CCL)\cite{Zhu_2016_CVPR} and Ensemble of Model Regression Trees (EMRT)\cite{DBLP:journals/corr/ZhuLLT15} developed head pose based and domain selective regressors respectively. \cite{Zhu_2016_CVPR} partitioned the optimization domain into multiple directions based on head pose and learned to combine the results of multiple domain regressors through composition estimator function. Similarly \cite{DBLP:journals/corr/ZhuLLT15} trained an ensemble of random forests to directly predict the locations of keypoints whereafter face alignment is achieved by aggregating the consensus of different models.\par 
Recently, methods using 3D models for face alignment have been proposed. PIFA\cite{pifa} by Jourabloo et al.  suggested a 3D approach that employed cascaded regression to predict the coefficients of 3D to 2D projection matrix and the base shape coefficients. Another recent work from Jourabloo et al.\cite{lfa3d} formulated the face alignment problem as a dense 3D model fitting problem, where the camera projection matrix and 3D shape parameters were estimated by a cascade of CNN-based regressors. However, \cite{Zhu_2016_CVPR} suggests that optimizing  the  base  shape  coefficients and projection is indirect and sub-optimal since smaller parameter errors are not necessarily equivalent to smaller alignment errors. 3DDFA\cite{DBLP:journals/corr/ZhuLLSL15} by Zhu et al. fitted a dense 3D face model to the image via CNN, where the depth data is modeled in a Z-Buffer.

Our work principally falls in the category of regression-based approaches and addresses the issue of adapting the cascade shape regression to unconstrained settings. KEPLER performs joint training on three fundamental tasks, namely, 3D pose, visibility of each keypoint and the location of keypoints, using only 2D color image. It also demonstrates that efficient joint training on the three tasks achieves superior performance. One of the closely related work is \cite{DBLP:conf/eccv/ZhangLLT14} where the authors used multi-tasking for many attributes, but did not leverage the intermediate features.
 
\section{KEPLER}

\begin{figure*}[htp!]
\centering
\includegraphics[width=14cm,height=6.5cm]{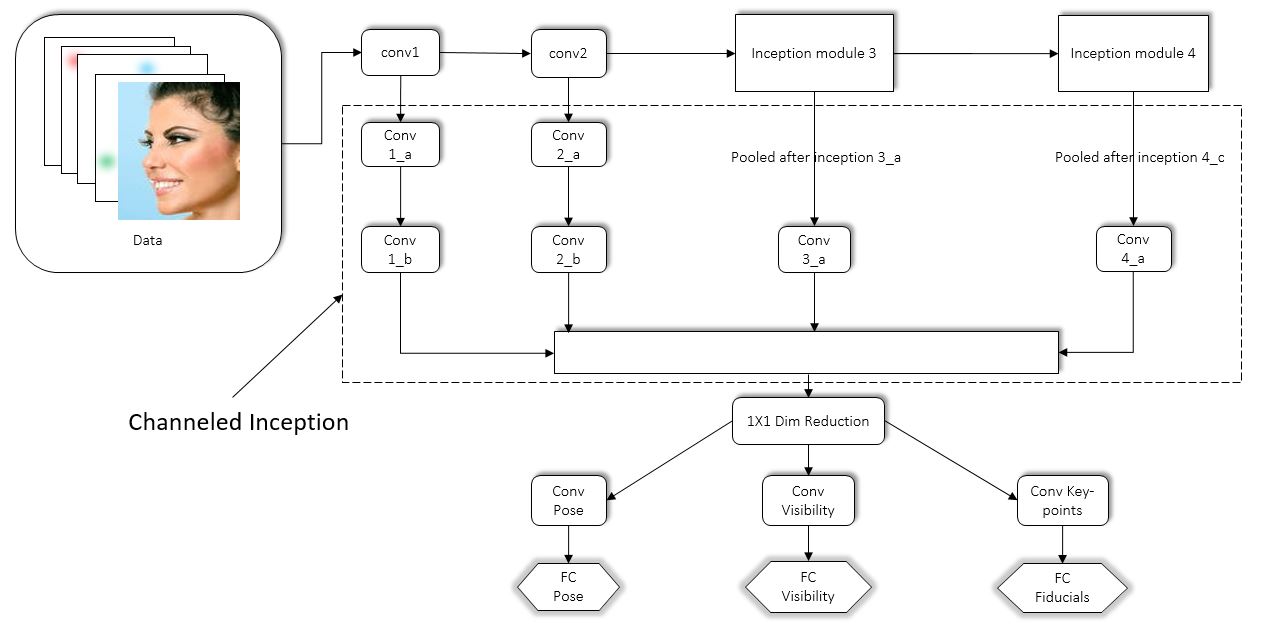}
\caption{The KEPLER network architecture. The dotted line shows the channeled inception network. The intermediate features are convolved and the responses are concatenated in a similar fashion as the inception module. Tasks such as pose are abstract and contained in deeper layers, however, the localization property is in the shallower layers. }
\label{fig:network_arch}
\end{figure*}

KEPLER is an iterative method which at its core consists of three modules. Figure \ref{fig:method} illustrates the basic building blocks of KEPLER. The first module is a rendering module \textit{h} which models the structure in an N-dimensional input space, with N being the maximum number of keypoints on a face. The current location of the keypoints are represented by the vector $\Vect{y_{t}}=\{y_{t}^{1}\ldots y_{t}^{N}\}$. The output of the rendering module is concatenated to the raw RGB input image $\Mat{I}$, along the third dimension which is then fed to the function \textit{f}.\\
The second module is the function \textit{f} which calculates the correction to be made at the next stage. The function \textit{f} is modeled by a convolution neural network whose architecture is described in  section \ref{netarch}. \\
The third module is the correction stage which adds the increments, predicted by \textit{f}, to the current locations. The output goes again into the rendering module \textit{h} which prepares the rendered data for the next iteration. The rendering function is not learned in this work, but represented by a 2D Gaussian with fixed variance and centered at current keypoint locations in each of the N channels. Finally, the Gaussian rendered images are stacked together with image $\Mat{I}$. Therefore the overall method can be summarized by the following set of equations. 
\begin{eqnarray}
\Vect{\delta_{t}} = f_{t}(\Mat{X_{t}},\Theta_t) \\
\Vect{y_{t+1}} = \Vect{y_{t}} + \Vect{\delta_{t}} \\
\Mat{X_{t+1}} = h(\Vect{y_{t+1}})
\end{eqnarray}
where \textit{f} is a function with learned parameters $\Theta_t$, predicting the increments $\Vect{\delta_{t}}$ . The prediction function \textit{f} is indexed by \textit{t} as it is trained separately for every iteration. In the first iteration, the function \textit{h} renders Gaussians at $y_0$, which is the mean shape. In this work we set $t=5$ iterations. We perform the last iteration only to take into effect the improper bounding box across different datasets (see Figure \ref{fig:sample}). The loss functions for each task is mentioned below.\\
\textbf{Keypoint localization}\\
Keypoint localization is the task of predicting the keypoints in a face. In this paper, we consider predicting the locations of \textit{N} = 21 keypoints on the face. With each point is associated the visibility of that point. The loss function for this task is given by 
\setlength{\belowdisplayskip}{5pt} 
\setlength{\abovedisplayskip}{5pt} 
\begin{equation}
L_{1}(\Vect{y},\Vect{g}) = \sum_{i = 1}^{N} v^{i}(y_{t}^{i} - g^{i})^{2}, 
\end{equation} 
where $y_{t}^{i}$ and $g^{i}$ are the predicted and the ground truth locations of the $i^{th}$ keypoint resprectively at time $t$. $v^{i}$ is the ground truth visibility associated with each keypoint. We discuss this loss function and its variant in section \ref{stage3}. \\
\textbf{Pose Prediction}\\
Pose prediction refers to the task of estimating the 3D pose of the face. We use the Euclidean loss function for pose prediction.
\setlength{\belowdisplayskip}{6pt} 
\setlength{\abovedisplayskip}{6pt} 
\begin{equation}
L_{2}(\Vect{p}_p,\Vect{g}_p) = (p_{yaw}-g_{yaw})^{2} + (p_{pitch}-g_{pitch})^{2} + (p_{roll}-g_{roll})^{2}
\end{equation} 
where $p$ stands for predicted and $g$ for the ground-truth.\\
\textbf{Visibility}\\
This task is associated with estimating the visibility of each keypoint.The number of keypoints visible on the face varies with pose. Hence, we use the Euclidean loss to estimate the visibility confidence of each point.
\setlength{\belowdisplayskip}{6pt} 
\setlength{\abovedisplayskip}{6pt} 
\begin{equation}
L_{3}(\Vect{v}_p,\Vect{v}_g) = \sum_{i = 1}^{N} (v_{p,i} - v_{g,i})^{2},
\end{equation} 
Therefore the net loss in the network is the weighted linear combination of the above loss functions.
\setlength{\belowdisplayskip}{10pt} 
\setlength{\abovedisplayskip}{10pt} 
\begin{equation}
L(p,g) = \lambda L_{1}(\Vect{y},\Vect{g}) + \mu L_{2}(\Vect{p}_p,\Vect{g}_p) + \nu L_{3}(\Vect{v}_p,\Vect{v}_g)
\end{equation}
where $\lambda$, $\mu$ and $\nu$ are the weight parameters suitably chosen depending on the iteration. 

\subsection{Network Architecture}  
\label{netarch}
\begin{figure}[h]
\centering
\includegraphics[width=2cm,height=2cm]{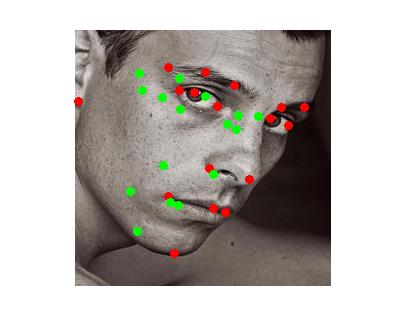}\includegraphics[width=2cm,height=2cm]{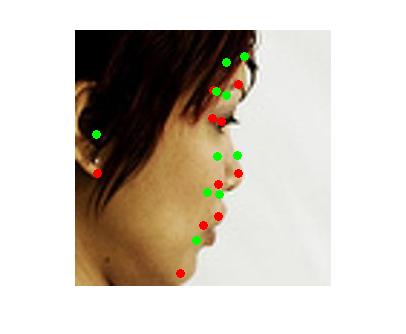}\includegraphics[width=2cm,height=2cm]{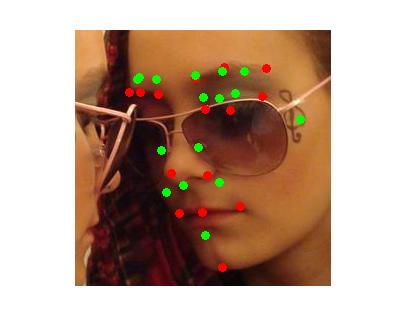}\includegraphics[width=2cm,height=2cm]{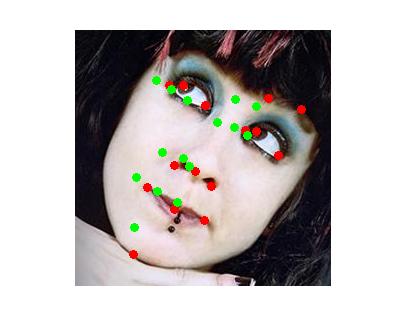}
\caption{\small Qualitative results of KEPLER after second stage. The green dots represent the predicted points after second stage. Red dots represent the ground truth. It can be seen that the visible  points have taken the shape of input face image.}
\label{fig:stage2}
\end{figure}
For the modeling function \textit{f} we design a unique ConvNet architecture based on GoogLenet\cite{c2} by pruning the inception network after inception\_4c. As PReLU has shown better performance in many vision tasks such as object recognition\cite{DBLP:journals/corr/HeZR015}, in this pruned network we first replace the ReLU non-linearity with PReLU.  We pool the intermediate features from the pruned GoogLenet. Then convolutions are performed from the output of each branch, and the output maps are concatenated similar to the inception module. We call this module the \textit{Channeled Inception} module. Since the output maps after $conv_{1}$ are larger in size, we first perform $4X4$ convolution and then again a $4X4$ convolution, both with the stride of 3 to finally match the dimension of the output to $7X7$. Similarly after $conv_{2}$ we first perform $4X4$ convolution and then $3X3$ convolution to match the output to $7X7$. The former uses a stride of 4 and the latter uses 2. The most n\"aive way of combining features is by concatenation. However, the concatenated output blob can be very high dimensional and hence we perform $1X1$ convolution for dimensionality reduction. This lets the network decide the weights to effectively combine the pooled features into lower dimension. It has been shown in \cite{DBLP:journals/corr/ZeilerF13} that adjacent layers are correlated and hence, we only pool features from alternate layers.\par
Next the network is trained on three tasks namely, pose, visibilities and the bounded error using ground truth. The joint training is helpful since it models the inherent relationship between visible number of points, pose and the amount of correction needed for a keypoint in particular pose. Choosing an architecture like GoogLenet is based on the fact that due to fewer number of parameters the training of GoogLenet is faster and adding to it batch normalization, even speeds up the training process. In order to further speed up the process we only use convolution layers till the last layer where we use a fully connected layer to get the final output.
The architecture of the whole network is shown in Figure \ref{fig:network_arch}.    

\subsection{Iteration 1 and 2: Constrained Training}

In this section, we explain the first stage training for keypoint estimation. The first stage is the most crucial one for face alignment. Since the network is trained from scratch, precautions have to be taken on what the network should learn. Directly learning the locations of keypoints from a network is difficult because when the network gets deeper it loses the localization capability. This is due to the fact that the outputs of the final convolution layers have a larger receptive field on the input image. We devise a strategy in which the corrections for the first two stages are bounded. 
Let us suppose the key-points are represented by their 2D coordinates $\Vect{y} :\{ \textit{y}^{i}\in \Re^{2}, i\in[1,\ldots,N]\}$ where \textit{N} is the number of keypoints and $y^i$ denotes the $i^{th}$ keypoint. The bounded corrections were calculated using (\ref{berr}) given below. 
\setlength{\belowdisplayskip}{6pt} 
\setlength{\abovedisplayskip}{6pt}
\begin{equation}
\label{berr}
\delta_{t}^{i}(g^i, y^{i}_{t}) = min(L,\|\Vect{u}\|).\hat{\Vect{u}}
\end{equation}
where \textit{L} denotes the bound of correction. $\Vect{u} = \Vect{g} - \Vect{y_{t}}$ and $\hat{\Vect{u}} = \frac{\Vect{u}}{\|\Vect{u}\|}$ represent the error vector and error unit vector respectively. In our experiments we set the bound $L$ to a maximum of 20 pixels. This simplifies the learning problem for the network in the first stage. According to this formulation, error correction for points for which the ground truth is far away, gets bounded by \textit{L}. The interesting property of this formulation is that in the first and second stage the network only learns the direction in which the points have to shift. This can be thought of as learning the direction of the unit error vector, to which the magnitude will be added later. In addition to just having keypoint location we also have access to facial 3D pose and the visibility of each point. One-shot prediction of the location of keypoints is difficult since the input space of the ConvNet is typically nonlinear. Also, learning small corrections should be easier, when the network is being trained for the first time. Hence, to impart prior knowledge to the network we jointly learn the pose and visibility of each point. The loss functions used for the three tasks are described in the previous section.\\
The function \textit{f} for second iteration is trained in a similar fashion with the weights initialized from the first iteration.

\subsection{Iteration 3: Variant of Euclidean loss}
\label{stage3}
We show the outputs of the network after the second stage of training in Figure \ref{fig:stage2}. Physical inspection of the outputs shows that for many of the faces, the network has already learned the magnitude and direction of the correction vector. However, there are misalignments in some images or in some keypoints in the images. But repeating the training methodology exactly as second iteration revealed that our architecture suffered from vanishing gradients. While back propagating the gradients, the loss is averaged over a batch and if there are few misalignments in a batch, there is very little gradient to be propagated. To maintain consistency we stick with the same architecture. Even though GoogLenet\cite{c2} claims to not have vanishing gradient problem, KEPLER faced it because of the dataset being small.

This motivated us to design a loss function that satisfies both of these conditions: on the one hand, the loss function should minimize the error between prediction and the ground truth; on the other hand, it should have sufficient gradients to be propagated to make the learning process reach global minima. Towards this end, we use the following loss function.
\setlength{\belowdisplayskip}{2pt} 
\setlength{\abovedisplayskip}{2pt}
\begin{eqnarray}
\label{loss}
L_{1}(\Vect{y},\Vect{g}) = \frac{1}{n}\left(\sum_{i = 1}^{N} v_{i}(y_{i} - g_{i})^{2} + \gamma \sum_{i = 1}^{N} v_{i}\mid y_{i} - g_{i}\mid\right) \\
\frac{\delta L_{1}(\Vect{y},\Vect{g})}{\delta \Vect{y}} = \frac{1}{n}\left(2 \sum_{i = 1}^{N} v_{i}(y_{i} - g_{i}) + \gamma \sum_{i = 1}^{N} v_{i}\frac{\mid y_i - g_i \mid}{y_i - g_i}\right)
\end{eqnarray}
where $\gamma$ is a parameter which controls the strength of the gradient and \textit{n} is the number of samples in a batch. We would like to emphasize that the additional term is not a regularizer as it is added to the objective function and does not directly regularize the weights. However, this is able to provide substantial gradients for the training of ConvNet.  \\
The representation function \textit{h} in this stage does not render any Gaussian in the channel for which the predicted visibility is below the threshold $\tau$. In this work we set this threshold $\tau$ to 0.03 and $\gamma$ to 0.2 obtained by cross validation. We do not constrain the amount of error corrections for the third stage training.
\subsection{Iteration 4: Hard sample mining} 
\label{stage4}
Recently, Kabkab et al.\cite{c4} suggested that by efficiently sampling the data one can make an optimal use of training set while training ConvNets leading to improved performance. \cite{c4} developed an online data sampling method based on a convex optimization formulation and showed how their formulation can make the classifier robust in class imbalanced problem. In our case, although after the third iteration, most of the images are aligned, they lack precision in local alignment. Inspired by \cite{c4}, we reuse the hard samples of the dataset to build a more robust keypoint localization system. \par
\begin{figure}[htp!]
\centering
\includegraphics[height=4cm,width = 8cm]{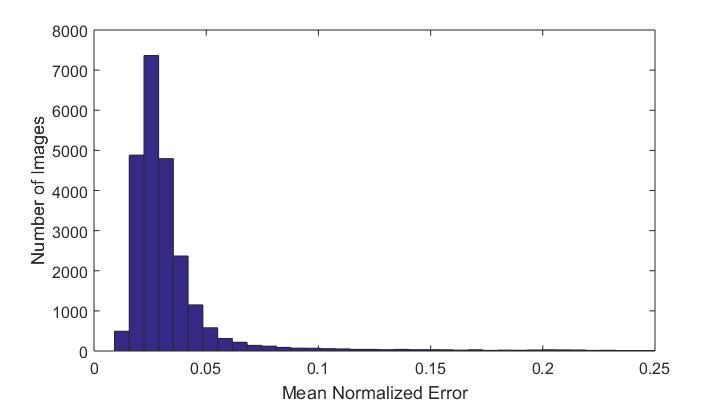}
\caption{\small Error Histogram of training samples after stage 3}
\label{fig:error_hist}
\end{figure}
Using the keypoints predicted after the third iteration, we plot the histogram (Fig.\ref{fig:error_hist}) of normalized mean error (NME), after calculating it for all the training samples. We denote the NME on x-axis around which the maximum number of samples are centered, as \textit{C}. In an ideal case, the value of \textit{C} should be low, implying that the average alignment error is less. Therefore, the objective of this stage is to lower the value of \textit{C} by hard sample mining. We select a threshold $\Delta$ ($0.03$ in our experiments), towards the right of \textit{C}, after which at least $30-40\%$ of the samples lie, as the threshold for hard samples. Using $\Delta$, we partition the dataset into two groups of hard and easy samples. We first select equal number of samples from both groups to form a batch which is then presented to ConvNet for training. This effectively results in reusing the hard samples. Then, to counter the group imbalance we finetune the network with entire dataset again with a lower learning rate. We use the loss function as in (\ref{loss}) with $\gamma = 0.1$ for this stage.
\subsection{Iteration 5: Local Error Correction}
\label{stage5}
There is a lot of inconsistency among the bounding boxes provided by different datasets. AFLW\cite{tugraz:icg:lrs:koestinger11b} provides larger bounding box annotations compared to AFW\cite{AFW_dataset_CVPR2012}. Regression-based alignment methods are dependent on the mean shape initialization, which is scaled to the bounding box size. Also it is impractical to come up with a heuristic which tries to determine compatible bounding boxes. Almost all the existing methods perform data augmentation by randomly perturbing the bounding boxes by some amount. However, it is not clear by how much the bounding boxes should be perturbed to obtain reasonably good bounding boxes during testing which is consistent with the dataset the network was trained on. We train our networks on a larger bounding box provided by AFLW. AFLW bounding boxes tend to be square and for almost all the images the nose tip appears at the center of the bounding box. This is a big limitation for the deployment of the system in real world scenarios. It is worthy to note that the previous four stages are trained on full images and hence produce global corrections. 
\begin{figure}[htp!]
\centering
\includegraphics[height=2.5cm,width = 8cm]{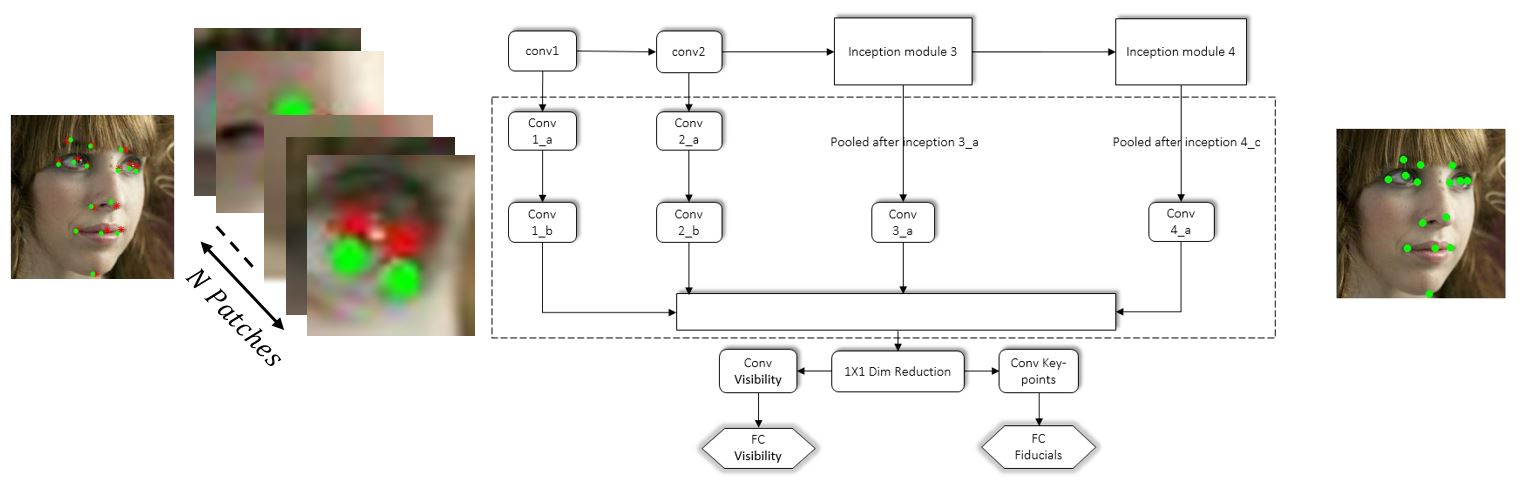}
\caption{\small Red dots in the left image represent the ground truth while green dots represent the predicted points after the fourth iteration. Local patches centered around predicted points are extracted and fed to the network. The network shown in Fig \ref{fig:network_arch} (see section \ref{stage5} for details) is trained on the task of local fiducial correction and visibility of fiducials inside the patch. The image on the right shows the predictions after local correction. }
\label{fig:stage5net}
\end{figure}
Our last stage of local correction is optional, which depends upon the test set and the bounding box annotations that it comes with. We train an exactly similar network as in Fig \ref{fig:network_arch}. but only for the tasks of predicting the visibility and corrections in the local patches. Predicting the pose with a local patch of say $WXW$ pixels is difficult which can lead the network to learn improper weights. We choose all the $N$ patches irrespective of the visibility factor. Learning visibility and corrections is important because we do not want the network to propagate any gradient if the point is invisible. We observe during experimentation that training the ConvNet on two tasks together achieves significantly better performance than when the network is trained only for the task of error correction. We again partition the dataset into easy and hard sample groups according to the strategy explained in the previous section. We finally finetune the network with the whole dataset with lower learning rate. 

\section{EXPERIMENTS AND COMPARISON}

\subsection{Datasets}

We select two challenging datasets with their most recent benchmarks.  \\
\textbf{\textit{In-the-wild datasets:}} To make the system robust for images in real life scenarios such as challenging shape variations and significant view changes, we select AFLW\cite{tugraz:icg:lrs:koestinger11b} for training and, AFLW and AFW\cite{AFW_dataset_CVPR2012} as the main test sets. \textbf{AFLW} contains $24,386$ in-the-wild faces (obtained from \textit{Flickr}) with head  pose ranging from $0\degree$ to $120\degree$ for yaw  and upto $90\degree$ for  pitch and roll with extremely challenging  shape  variations  and  deformations.  Along with this AFLW also demonstrates external-object occlusion. There are a total of 21\% invisible landmarks caused by occlusion, larger than 13\% on COFW\cite{10.1109/ICCV.2013.191} where only internal object-occlusion is exhibited. 
In addition, one important point to note is that COFW also provides the annotations for the invisible landmarks while in the case of AFLW the invisble landmarks are absent. \textbf{AFW} is a popular benchmark for the evaluation of face alignment algorithms. AFW contains 468 in-the-wild faces (obtained from Flickr) with yaw degree up to $90\degree$ . The images are well diverse in terms of pose, expression and illumination. The number of visible points also varies depending on the image, but the location of occluded points are to be predicted as well.\\
AFLW provides at most 21 points for each face. It  excludes coordinates for invisible landmarks, which we consider to be the best, because there is no way of correctly knowing the exact location of those points. In many cases such invisible points are mostly hallucinated and annotated thereafter. \\   \\
\textbf{{Testing Protocols:}} \\ \\
\textbf{(I)AFLW-PIFA:} We follow the protocol used in PIFA\cite{pifa}. We randomly select $23,386$ images for training and the remaining $1,000$ for testing. We divide the testing images in three groups as done in \cite{pifa}: $[0\degree,30\degree]$, $[30\degree,60\degree]$ and $[60\degree,90\degree]$ where the number of images in each group are taken to be equal.\\ 
\textbf{(II)AFLW-Full:} We  also test on the full test set of AFLW of sample size $1,000$. \\ 
\textbf{(III)AFLW-All variants:} In the next experiment, to have more rigorous analysis, we perform the test on all variants of images from (I) above. To create all variants images, we first rotate the whole images from (I) at angles of $15\degree$,$30\degree$,$45\degree$ and $60\degree$. We do the same with the horizontally flipped version of these images. We then rotate the bounding box coordinates and the key-points also at the same angles and crop the faces. This is done for all the images following the AFLW-PIFA protocol. One important effect of this rotation is that some of the images have smaller face compared to others due to rotated bounding box. This experiment tests the robustness of the algorithm on faces of different effective sizes and orientations.\\ 
\textbf{(IV)AFW:} We only use AFW for testing purposes. We follow the protocol as stated in \cite{AFW_dataset_CVPR2012}. AFW provides 468 images in total, out of which 341 faces have height greater than 150 pixels. We only evaluate on those 341 images following the protocol of \cite{AFW_dataset_CVPR2012}.\\ \\
\textbf{Evaluation metric:}  Following  most  previous  works, we obtain the error for each test sample via averaging normalized errors for all annotated landmarks. We demonstrate our results with mean error over all samples, or via Cumulative Error Distribution (CED) curve. For pose, we evaluate on continuous pose predictions as well as their discretized versions rounded to nearest $15\degree$. We report the continuous mean absolute error for the AFLW testset and plot the Cumulative Error Distribution curve for AFW dataset. All the experiments including training and testing were performed using the Caffe\cite{jia2014caffe} framework and two Nvidia TITAN-X GPUs. Our method can process upto 3-4 frames per second, which can be higher in batch mode. 

\begin{table}[h]
\centering
\begin{tabular}{lll}
\hline
 & {{\textbf{\textit{AFLW}}}} & {{\textbf{\textit{AFW}}}} \\ 
\hline
{{\textbf{Method}}} & {{\textbf{NME}}}  & {{\textbf{NME}}}\\ \hline
TSPM \cite{AFW_dataset_CVPR2012}   & -     & 11.09   \\           
CDM \cite{Xiang_iccv_2013}         & 12.44    & 9.13          \\
RCPR \cite{10.1109/ICCV.2013.191}        & 7.85     &-          \\
ESR \cite{DBLP:journals/ijcv/CaoWWS14}         & 8.24    & -          \\
PIFA \cite{pifa}        & 6.8          & 8.61      \\
3DDFA \cite{DBLP:journals/corr/ZhuLLSL15}       & 5.32      &-         \\
LPFA-3D \cite{lfa3d}     & 4.72  & 7.43             \\
EMRT \cite{DBLP:journals/corr/ZhuLLT15}     & 4.01    & 3.55            \\ 
CCL \cite{Zhu_2016_CVPR}         & 5.85       & 2.45        \\ 
Rec Enc-Dec\cite{recdec}  & \textgreater6        & -        \\ \hline 
\textbf{KEPLER}       & \textbf{2.98}   & 3.01    \\ \hline
\end{tabular}
\caption{\small {Comparison of KEPLER with other state of the art methods. NME stands for normalized mean error. For AFLW, numbers for other methods are taken from respective papers following the PIFA protocol. For AFW, numbers are taken from respective works published following the protocol of \cite{AFW_dataset_CVPR2012}.}}
\label{aflw_table}
\end{table}

\begin{figure}[htp!]
\centering
\includegraphics[height=5.5cm,width=8cm]{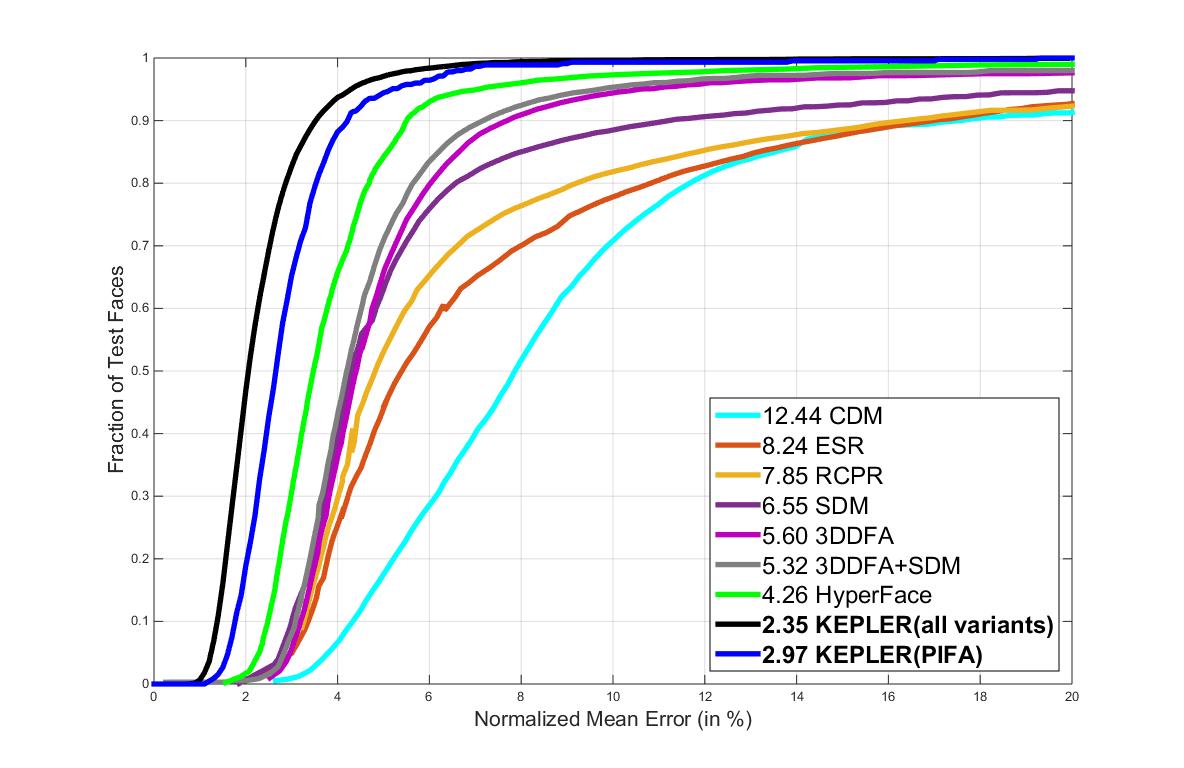}
\caption{\small Cumulative error distribution curves for landmark localization on
the AFLW dataset. The numbers in the legend are the average normalized mean error normalized by the face size.}
\label{aflw_res}
\end{figure}

\subsection{Results}
\label{com}
\begin{figure}[htp!]
\centering
\includegraphics[height=5.5cm,width=8cm]{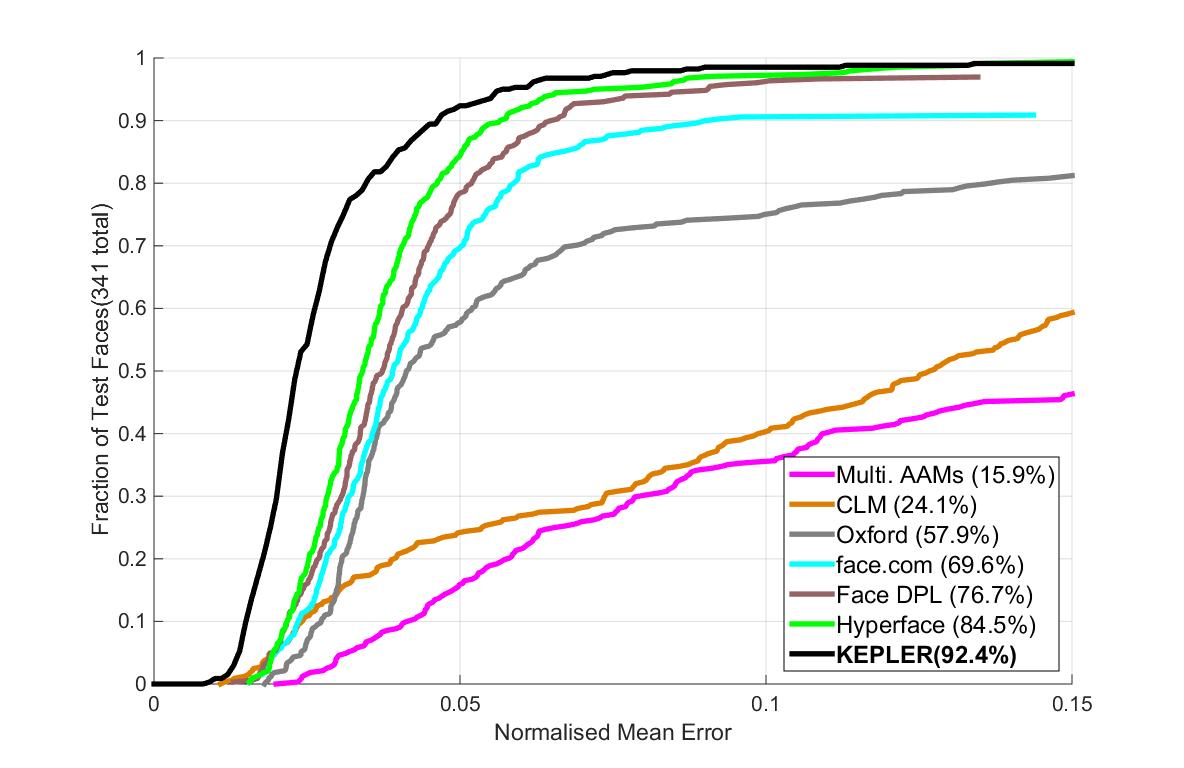}
\caption{\small Cumulative error distribution curves for landmark localization on
the AFW dataset. The numbers in the legend are the fraction of testing
faces that have average error below (5\%) of the face size.}
\label{afw_res}
\end{figure}

\begin{figure}[htp!]
\centering
\includegraphics[height=5.5cm,width=8cm]{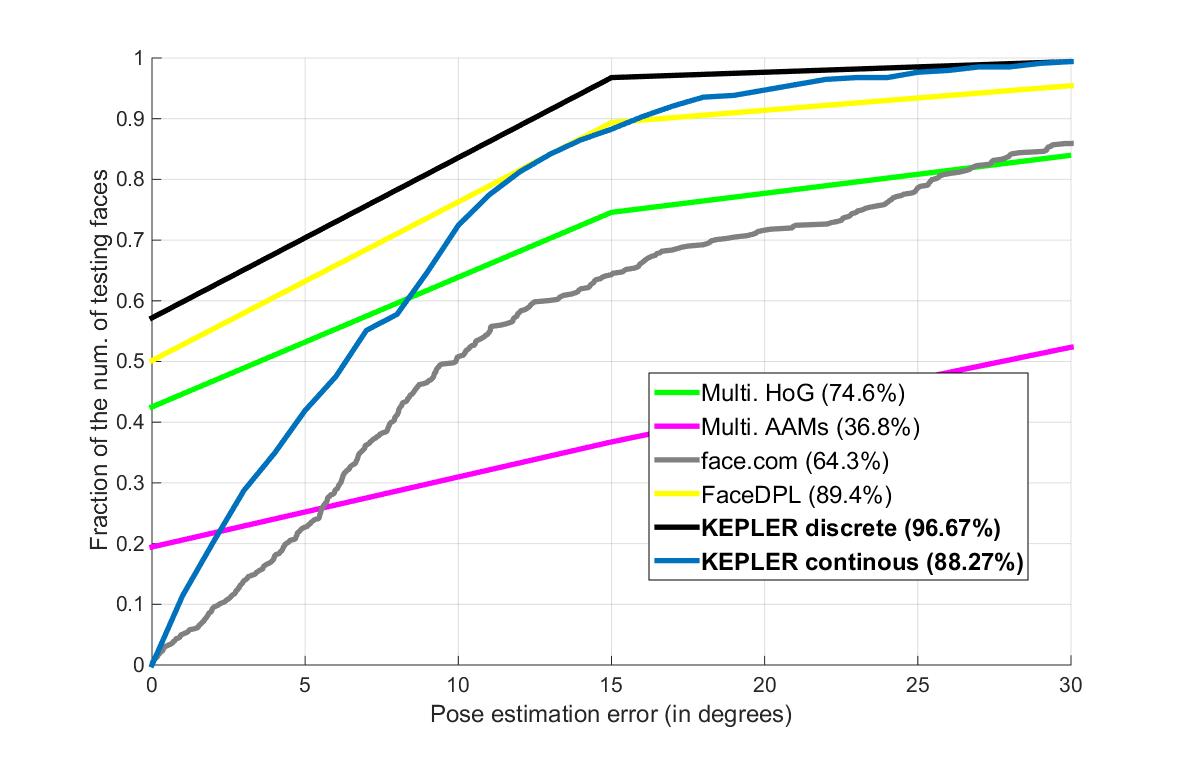}
\caption{\small Cumulative error distribution curves for pose estimation on AFW
dataset. The numbers in the legend are the percentage of faces that are
labeled within $\pm15\degree$ error tolerance}
\label{afw_pose}
\end{figure}

Table \ref{aflw_table} compares the performance of KEPLER compared to other existing methods. Table \ref{kepler_summary} summarises the performance of KEPLER under different protocols of AFLW testset. Table \ref{aflw_pose} shows the mean error in degrees, in estimating the 3D pose of a face image. Figures \ref{aflw_res} and \ref{afw_res} show the cumulative error distribution in predicting keypoints on the AFLW and AFW test sets. Figure \ref{afw_pose} shows the cumulative error distribution in pose estimation on AFW. 

\textbf{Comparison with CCL\cite{Zhu_2016_CVPR}}: It is clear from the tables that KEPLER outperforms all state of the art methods on the AFLW dataset. It also outperforms all state of the art methods except CCL\cite{Zhu_2016_CVPR} on the AFW datatset. Visual inspection of our results suggests that KEPLER is a little farther from ground truth on invisible points. We note that CCL\cite{Zhu_2016_CVPR} manually annotates the AFLW dataset with 19 landmarks along with the invisible landmarks, leaving the earpoints. In our experiments we prefer to use the dataset as provided by AFLW\cite{tugraz:icg:lrs:koestinger11b}, although we believe that CCL-kind of reannotation may boost the performance(since during AFW evaluation the locations of occluded points also need to be predicted). In KEPLER there is no loss propagated for the invisible points. We believe that training KEPLER on the revised annotation by \cite{Zhu_2016_CVPR} would make the prediction of occluded points more precise.  \\ 
\begin{table}[h]
\centering
\resizebox{\columnwidth}{!}{%
\begin{tabular}{lllll}
\hline
{\textbf{Method}} & {\textbf{AFLW-PIFA}} & {\textbf{AFLW-FULL}} & {\textbf{AFLW-Allvariants}} & {\textbf{AFW}}\\ \hline
\textbf{KEPLER}   & 2.98 & 2.90 & 2.35 & 3.01     \\ \hline
\end{tabular}%
}
\caption{\small Summary of performance on different protocols of AFLW and AFW by KEPLER.}
\label{kepler_summary}
\end{table}

\begin{figure*}[htp!]
 \centering
\includegraphics[width=2cm,height=2cm]{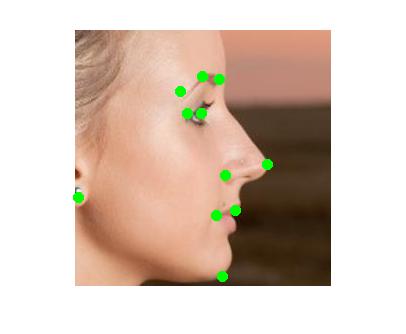}\includegraphics[width=2cm,height=2cm]{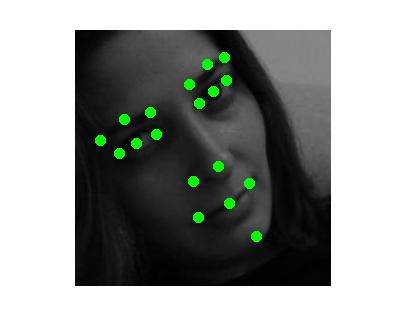}\includegraphics[width=2cm,height=2cm]{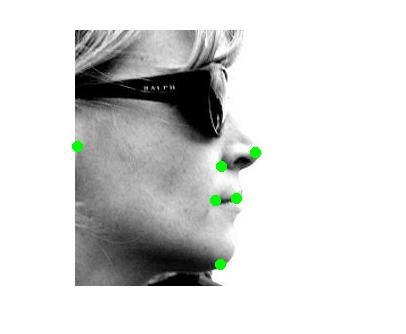}\includegraphics[width=2cm,height=2cm]{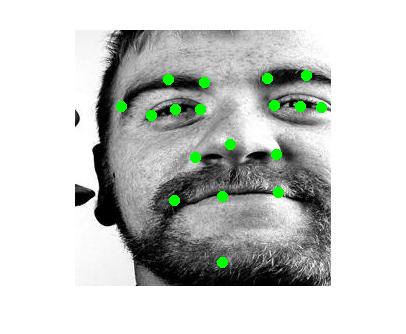}\includegraphics[width=2cm,height=2cm]{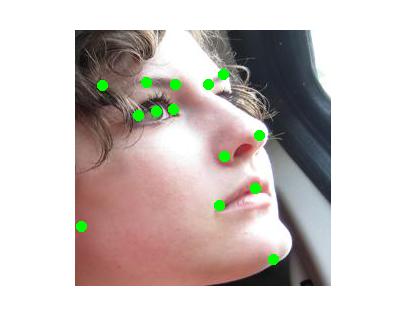}\includegraphics[width=2cm,height=2cm]{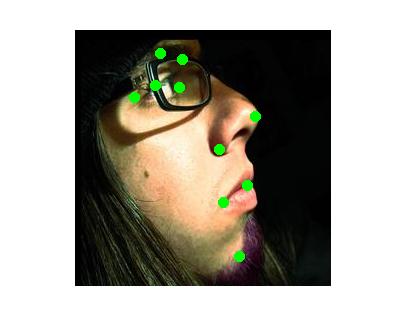}\includegraphics[width=2cm,height=2cm]{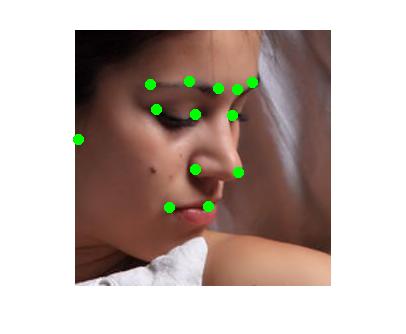}\includegraphics[width=2cm,height=2cm]{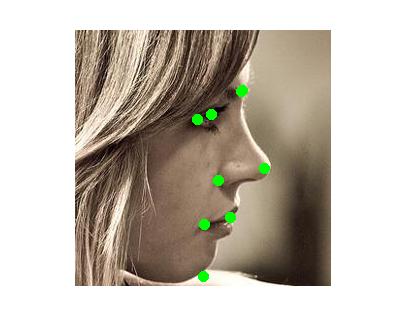}\includegraphics[width=2cm,height=2cm]{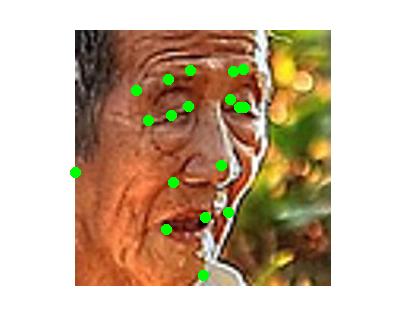}\\
\includegraphics[width=2cm,height=2cm]{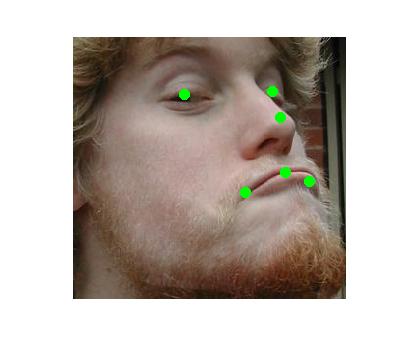}\includegraphics[width=2cm,height=2cm]{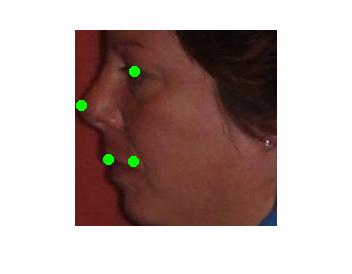}\includegraphics[width=2cm,height=2cm]{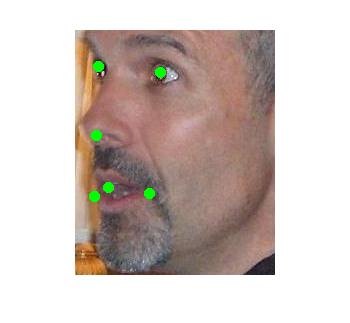}\includegraphics[width=2cm,height=2cm]{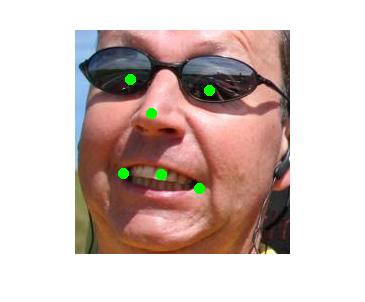}\includegraphics[width=2cm,height=2cm]{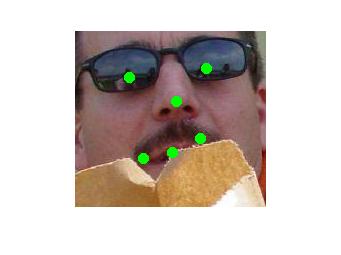}\includegraphics[width=2cm,height=2cm]{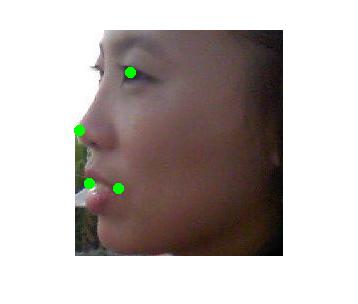}\includegraphics[width=2cm,height=2cm]{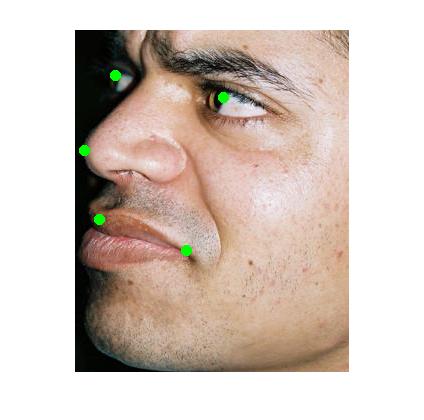}\includegraphics[width=2cm,height=2cm]{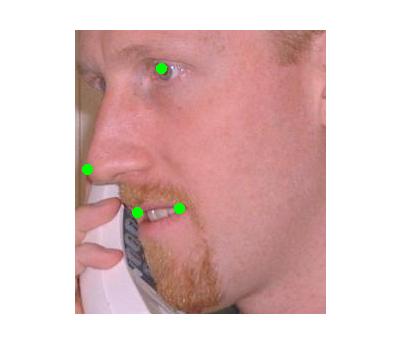}\includegraphics[width=2.25cm,height=2.1cm]{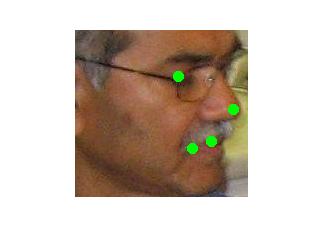}\\
\includegraphics[width=2cm,height=2cm]{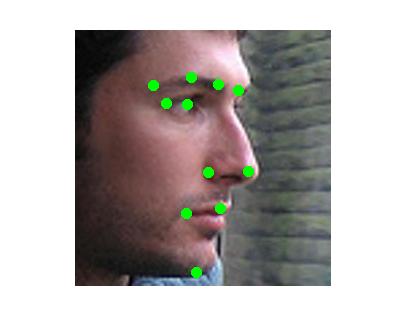}\includegraphics[width=2cm,height=2cm]{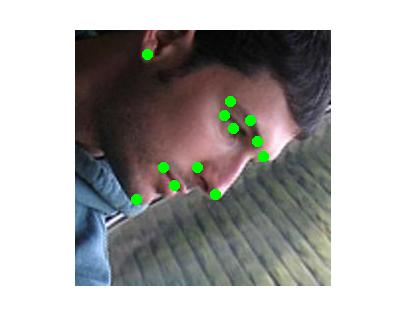}\includegraphics[width=2cm,height=2cm]{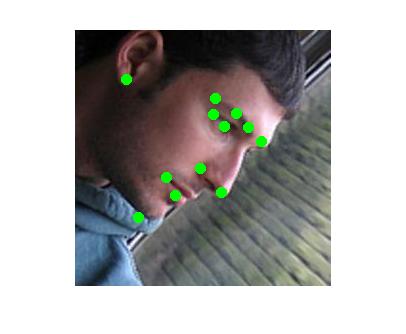}\includegraphics[width=2cm,height=2cm]{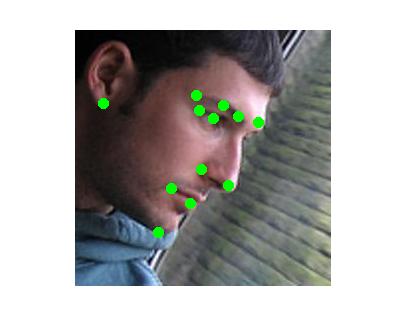}\includegraphics[width=2cm,height=2cm]{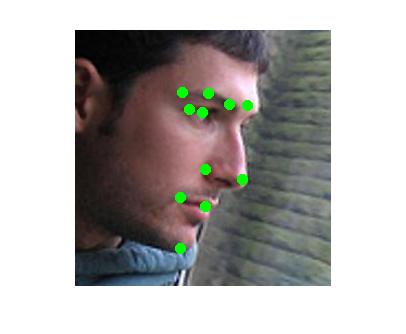}\includegraphics[width=2cm,height=2cm]{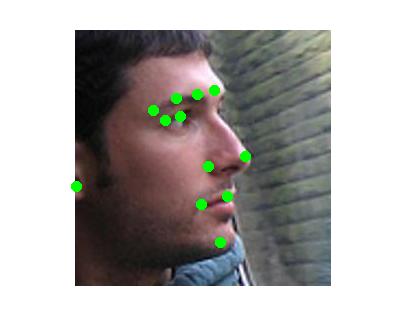}\includegraphics[width=2cm,height=2cm]{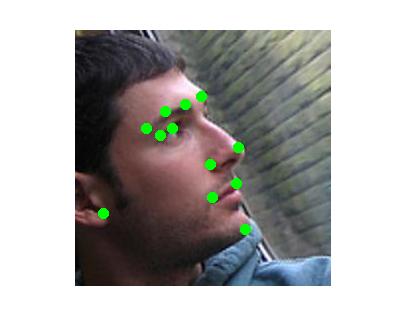}\includegraphics[width=2cm,height=2cm]{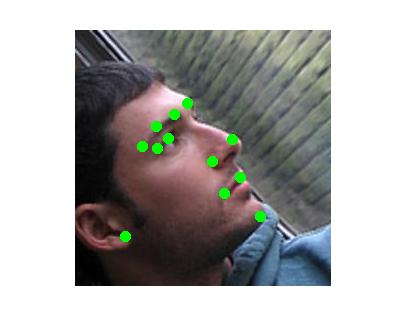}\includegraphics[width=2cm,height=2cm]{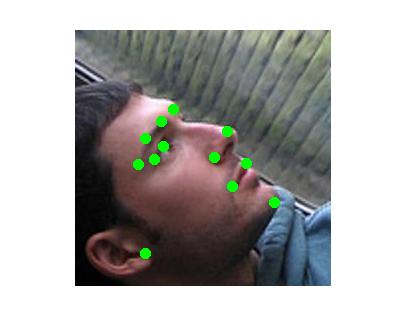}\\
\includegraphics[width=2cm,height=2cm]{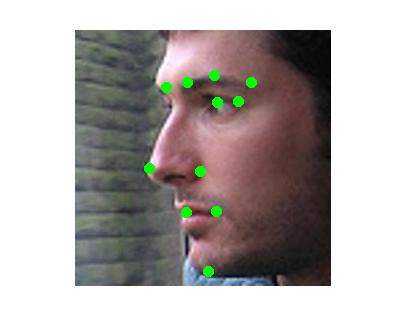}\includegraphics[width=2cm,height=2cm]{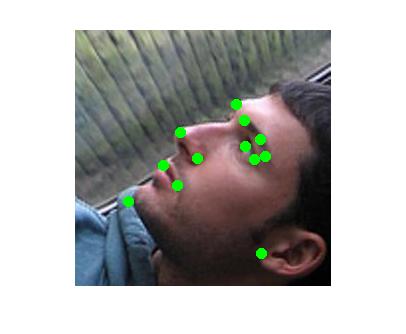}\includegraphics[width=2cm,height=2cm]{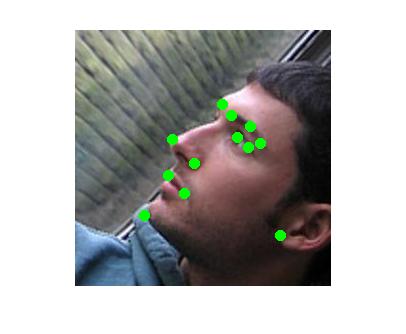}\includegraphics[width=2cm,height=2cm]{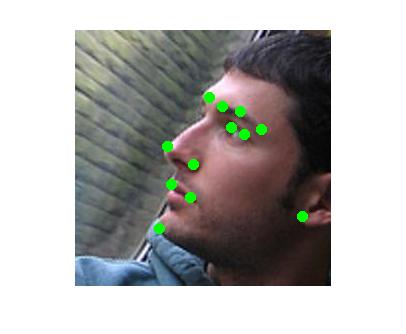}\includegraphics[width=2cm,height=2cm]{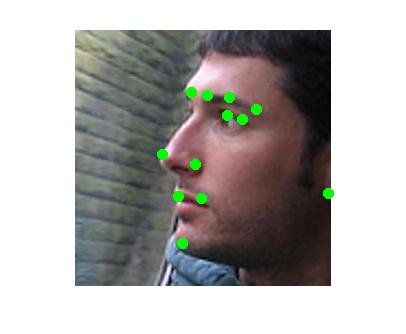}\includegraphics[width=2cm,height=2cm]{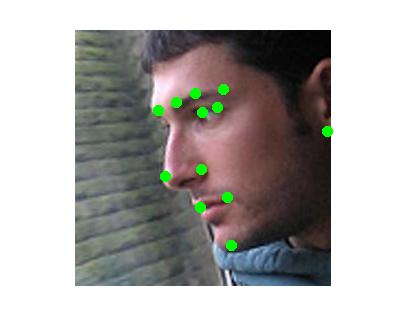}\includegraphics[width=2cm,height=2cm]{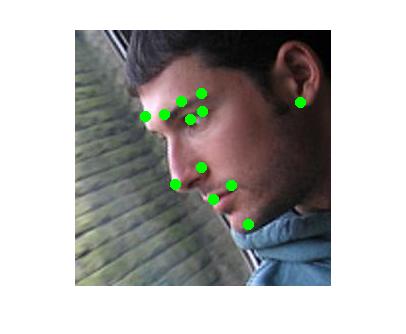}\includegraphics[width=2cm,height=2cm]{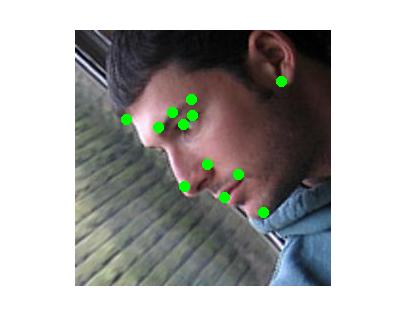}\includegraphics[width=2cm,height=2cm]{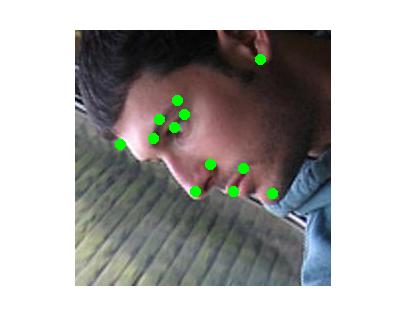}
\caption{\small Qualitative results of KEPLER after last stage. The green dots represent the final predicted points after fifth iteration. First row are the test samples from AFLW. Second row shows the samples from AFW dataset. The last two rows are the results of KEPLER after last stage from AFLW testset for all variants protocol. The green dots represent the final predicted points after fifth iteration.}
\label{fig:allstage5}
\end{figure*}

\begin{table}[h]
\centering
\resizebox{\columnwidth}{!}{%
\begin{tabular}{c|cccc|c}
\cline{1-6}
&  \multicolumn{4}{ c| }{\textbf{AFLW}} &\multicolumn{1}{c}{\textbf{AFW}} \\ \cline{2-6}
\textbf{Method} &  \textbf{Yaw} & \textbf{Pitch} & \textbf{Roll} & \textbf{MAE} & $\bf{Accuracy(\leq15\degree)}$ \\ \cline{1-6}
\multicolumn{1}{c|}{Random Forest\cite{Valle2016}} & 
\multicolumn{1}{c|} - & - & - & 12.26\degree & 83.54\%     \\ \cline{1-6}
\multicolumn{1}{c|}{\textbf{KEPLER}} &
\multicolumn{1}{c|} {\bf{6.45\degree}} & \textbf{5.85\degree} & \textbf{8.75\degree} & \textbf{6.45\degree} & \textbf{96.67\%}  \\ \cline{1-6}
\end{tabular}%
}
\caption{\small Comparison of Mean error in 3D pose estimation by KEPLER on AFLW testset. For AFLW \cite{Valle2016} only compares mean average error in Yaw. For AFW we we compare the percentage of images for which error is less than 15\degree.}
\label{aflw_pose}
\end{table}
We also verify our claim that iteration 5 is optional and only required for transferring the algorithm to other datasets with different bounding box annotations. To support our claim we calculate the normalized mean error after iteration 4 for both datasets and compare with the error obtained after iteration 5. The error after iteration 4 for AFLW testset was 0.0369 (which is already lower than all existing works) and after fifth iteration it was 0.0299, bringing the performance up by $18\%$. On the other hand the improvement in AFW (whose bounding box annotation is different from AFLW) was close to $60\%$. The error after iteration 4 on AFW dataset was 0.0757 which decreases to  0.0301 after fifth iteration. \\
We demonstrate some qualitative results from AFLW and AFW test sets in Figure \ref{fig:allstage5}.

\section{Conclusions and Future Works}
In this work we show that by efficiently capturing the structure of face through additional channels, we can obtain precise keypoint localization on unconstrained faces. We propose a novel \textit{Channeled Inception} deep network which pools features from intermediate layers and combines them in the same manner to the Inception module. We show how cascade regressors can outperform other recently developed works. As a byproduct of KEPLER, 3D facial pose is also generated which can be used for other tasks such as pose dependent verification methods, 3D model generation and many others. In conclusion, KEPLER demonstrates that by improved initialization and multitask training, cascade regressors outperforms state of the art methods not only in predicting the keypoints but also for head pose estimation.

\section{Acknowledgment}
This research is based upon work supported by the Office of the Director of National Intelligence (ODNI), Intelligence Advanced Research Projects Activity (IARPA), via IARPA R\&D Contract No. 2014-14071600012. The views and conclusions contained herein are those of the authors and should not be interpreted as necessarily representing the official policies or endorsements, either expressed or implied, of the ODNI, IARPA, or the U.S. Government. The U.S. Government is authorized to reproduce and distribute reprints for Governmental purposes notwithstanding any copyright annotation
thereon.

{\small
\bibliographystyle{ieee}
\bibliography{FD}
}

\end{document}